\pdfoutput=1

\documentclass[11pt]{article}

\usepackage[final]{acl}

\usepackage{times}
\usepackage{latexsym}

\usepackage[T1]{fontenc}

\usepackage[utf8]{inputenc}

\usepackage{microtype}

\usepackage{inconsolata}

\usepackage{graphicx}

\usepackage{enumitem}

\usepackage{amsmath}
\usepackage{amssymb}
\usepackage{multicol}

\usepackage{booktabs}
\usepackage{multirow}
\usepackage{dcolumn}
\usepackage{threeparttable}
\usepackage{makecell}
\usepackage{tabularx}
\newcolumntype{d}{D{.}{.}{-1}}
\newcolumntype{Y}{>{\centering\arraybackslash}X}

\usepackage{listings}
\lstdefinestyle{mystyle}{
  backgroundcolor=\color{gray!10},
  basicstyle=\ttfamily\small,
  frame=single, framesep=6pt, framerule=0.5pt, rulecolor=\color{black!30},
  breaklines=true, breakatwhitespace=true, breakindent=0pt,
  showstringspaces=false, columns=flexible, captionpos=b,
  numberstyle=\tiny\color{gray},
  xleftmargin=10pt, xrightmargin=10pt, tabsize=4, extendedchars=true
}
\title{Quantifying Data Contamination in Psychometric Evaluations of LLMs}

\author{Jongwook Han$^*$ ~~ Woojung Song$^*$ ~~ Jonggeun Lee$^*$ ~~ Yohan Jo$^{\dag}$ \\
  Graduate School of Data Science, Seoul National University \\
  \texttt{\{johnhan00,opusdeisong,jonggeun.lee,yohan.jo\}@snu.ac.kr} \\
  }

\begin{document}
\maketitle
\def\thefootnote{\fnsymbol{footnote}}
\footnotetext[1]{Equal contribution.}
\footnotetext[2]{Corresponding author.}
\def\thefootnote{\arabic{footnote}}
\begin{abstract}
Recent studies apply psychometric questionnaires to Large Language Models (LLMs) to assess high-level psychological constructs such as values, personality, moral foundations, and dark traits. Although prior work has raised concerns about possible data contamination from psychometric inventories, which may threaten the reliability of such evaluations, there has been no systematic attempt to quantify the extent of this contamination. To address this gap, we propose a framework to systematically measure data contamination in psychometric evaluations of LLMs, evaluating three aspects: (1) item memorization, (2) evaluation memorization, and (3) target score matching. Applying this framework to 21 models from major families and four widely used psychometric inventories, we provide evidence that popular inventories such as the Big Five Inventory (BFI-44) and Portrait Values Questionnaire (PVQ-40) exhibit strong contamination, where models not only memorize items but can also adjust their responses to achieve specific target scores.\footnote{Our code is available at \url{https://github.com/holi-lab/psychometric-contamination}.}
\end{abstract}

\section{Introduction\label{sec:introduction}}
As Large Language Models (LLMs) are increasingly deployed in real-world applications, evaluating their characteristics and behaviors has become an important question in natural language processing. Recent work addresses this question through applying established psychometric frameworks such as the Big Five Inventory~\citep{john1991big} and the Portrait Values Questionnaire (PVQ)~\citep{schwartz2012overview} to assess personality traits and value orientations~\citep{miotto-etal-2022-gpt, huang-etal-2024-reliability}.

However, several studies have raised concerns that relying on established questionnaires entails the risk of data contamination, as LLMs may have encountered psychometric items or scoring procedures during pretraining~\citep{insights, lee-etal-2025-llms, ye2025gpv, han-etal-2025-value}. As a result, the psychometric test results of LLMs may reflect prior exposure in training data rather than their genuine characteristics~\citep{hagendorff2024machinepsychology}. Moreover, such contamination can induce models to reproduce responses aligned with the expected personality or value profiles described in scientific literature, rather than revealing their intrinsic tendencies~\citep{miotto-etal-2022-gpt}.

Despite these concerns, no prior work has further examined the issue in depth, mainly due to the lack of a systematic method for quantifying such contamination. Motivated by this, we propose measures specifically designed for the psychometric evaluation of LLMs.

\begin{figure}[t]
    \centering
    \includegraphics[width=1.0\linewidth]{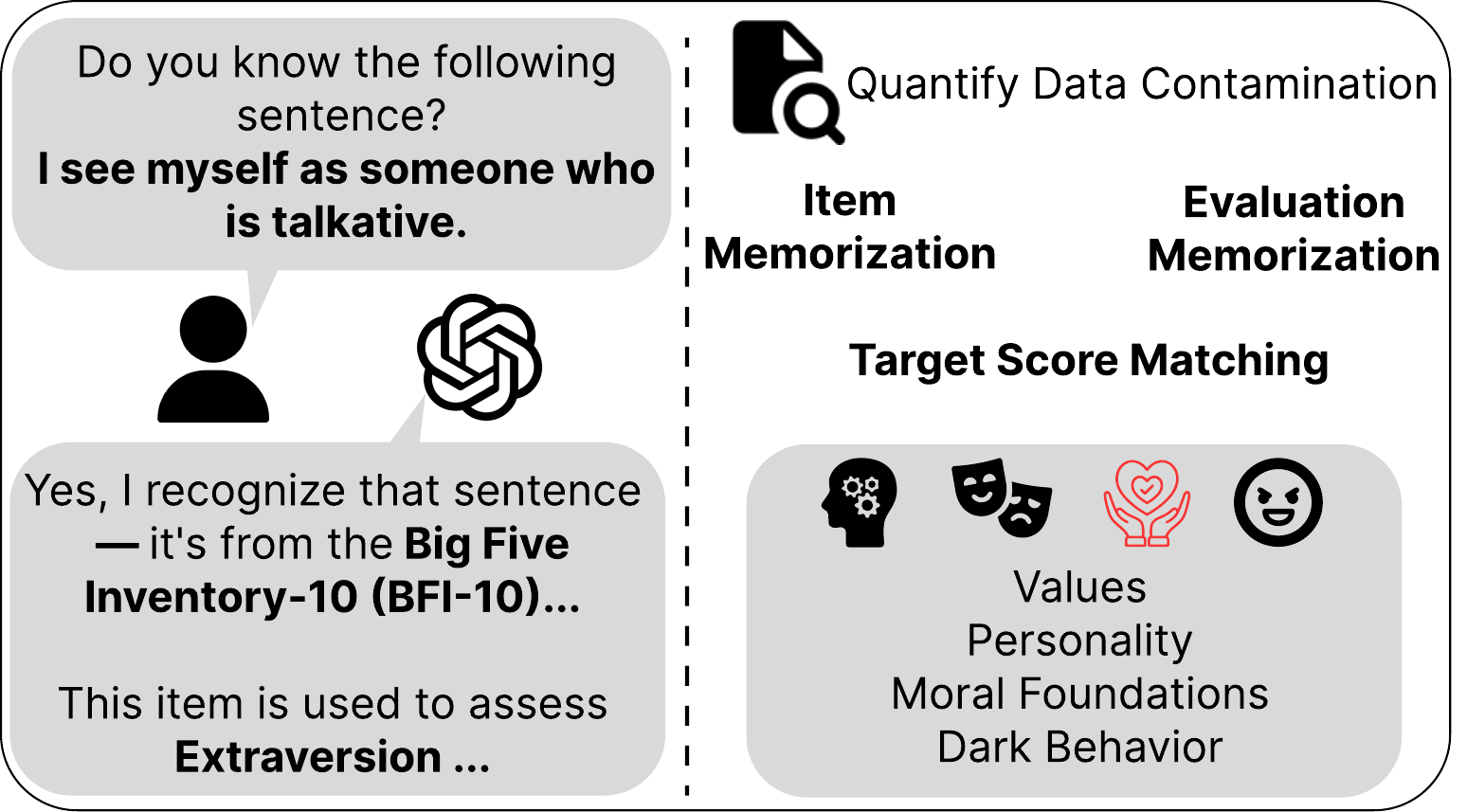}
    \caption{(Left) LLMs can recognize psychometric inventory items. (Right) Our framework quantifies data contamination across three aspects: \textit{item memorization}, \textit{evaluation memorization}, and \textit{target score matching}.}
    \label{fig:figure1}
\end{figure}

Psychometric inventories differ fundamentally from standard NLP benchmarks. Inventory items are explicitly designed to measure psychological constructs rather than solve task-specific problems, and each inventory comes with its own response scales and scoring procedures (e.g., reverse-coded items). Because of these characteristics, contamination may arise in multiple forms, including memorization of item wording, recognition of construct associations, and familiarity with scoring rules. This calls for a methodology specifically tailored to psychometric evaluation.

In this work, we propose a framework that systematically quantifies data contamination in psychometric evaluations of LLMs. The framework covers three aspects: (1) item memorization (direct exposure to inventory items), (2) evaluation memorization (familiarity with scoring protocols), and (3) target score matching (the ability to steer toward a desired score). We apply this framework to multiple model families and widely used inventories, including the Big Five Inventory (BFI-44), PVQ, Moral Foundations Questionnaire (MFQ)~\citep{MFQ-30}, and the Short Dark Triad (SD-3)~\citep{sd3}. We demonstrate that, as researchers have been concerned, a majority of LLMs are contaminated with these inventories. Our work paves the way for more systematic investigations into whether these models manipulate such content in psychometric tests in future research.

\section{Quantifying Data Contamination\label{sec:quantify}}

\paragraph{Motivation.}
As illustrated in Figure~\ref{fig:figure1}, GPT-4o is able to identify that a given item belongs to the Big Five Inventory (BFI-10)~\citep{BFI-10} and even knows that the item is related to the ``Extraversion'' dimension. This suggests that the model may already possess prior knowledge of BFI-10. However, to our knowledge, no existing research has systematically measured how much LLMs are familiar with specific psychometric inventories. To address this gap, we focus on three aspects of potential contamination.
\begin{itemize}[noitemsep, topsep=0pt,itemsep=0pt,partopsep=0pt,parsep=0pt, leftmargin=1.5em]

    \item \textbf{Item Memorization}: whether LLMs know the items of psychometric inventories.
    \item \textbf{Evaluation Memorization}: whether LLMs understand the scoring procedures of the inventory items.
    \item \textbf{Target Score Matching}: whether LLMs can adjust their responses to match a specified target score.

\end{itemize}

\subsection{Item Memorization\label{sec:item_memorization}}
We define \textit{item memorization} as an LLM's memorization of items from a psychometric inventory. To evaluate this, we design two tasks: \textit{Semantic Memorization} and \textit{Key Information Memorization}.

\paragraph{Semantic Memorization.} 

To analyze whether LLMs have prior information about psychometric items, we evaluate their ability to reconstruct item content when provided with the inventory name and item index. We prompt models to generate the corresponding item text. Then we filter out refusal responses using GPT-5.2-mini as an automated filter. For the remaining responses, we compute the embedding similarity between the original and the generated item text. We use OpenAI's text-embedding-3-large model and compute cosine similarity. We focus on semantic similarity rather than exact text overlap, as models may paraphrase item content.

\paragraph{Key Information Memorization.}
We also evaluate whether LLMs memorize the key semantic content of inventory items. For each item $i_n$, we mask the most informative keyword (e.g., in the first item of BFI-10, ``I see myself as someone who is reserved'', the word ``reserved'' is masked). Keywords were annotated by a psychology expert, and the model is prompted with the masked item to generate the missing keyword. To validate the the reliability of the keyword annotations, we collected three additional human annotations and computed the Fleiss' Kappa among the four annotators. The resulting agreement scores were 0.9187 for BFI-44, 0.7893 for MFQ, 0.8168 for PVQ-40, and 0.7548 for SD3. These results indicate that the inter-annotator agreement is consistently high across inventories, suggesting that the masked-keyword annotations are reliable.

We compute the success rate as $\textstyle\frac{1}{N}\sum_{n=1}^{N}\mathbf{I}(k_n = \hat{k}_n)$, where $N$ is the number of items, $k_n$ is the gold keyword, $\hat{k}_n$ is the model prediction, and $\mathbf{I}(\cdot)$ is the indicator function. A higher success rate indicates stronger memorization of the core meaning of inventory items.

\subsection{Evaluation Memorization}
Even if an LLM memorizes an item, this does not imply that the model understands what the item measures or how it should be evaluated. We assess this through two tasks: \textit{Item-Dimension Mapping} and 
\textit{Option-Score Mapping}.

\paragraph{Item-Dimension Mapping.}
We prompt LLMs with each item from an inventory along with the full set of dimensions the inventory measures (e.g., Openness, Conscientiousness, Extraversion, Agreeableness, Neuroticism for BFI-10). The models are then asked to identify which dimension each item corresponds to. We use the averaged F1-score across all dimensions:

\[
\operatorname{F1} = \frac{1}{D} \sum_{d=1}^{D} 
\frac{2 \cdot \operatorname{Precision}_d \cdot \operatorname{Recall}_d}{\operatorname{Precision}_d + \operatorname{Recall}_d},
\]
where $D$ is the total number of dimensions, and $\operatorname{Precision}_d$ and $\operatorname{Recall}_d$ 
are computed with respect to dimension $d$. A high F1-score indicates that the model knows the dimension each item measures.

\paragraph{Option-Score Mapping.} Beyond recognizing item-dimension relations, we also evaluate whether models understand how response options correspond to numeric scores under inventory-specific scoring rules, including reverse-coded items.

To evaluate this capability, we prompt each model with an item, its response options, and the target dimension, asking it to output the numeric scores defined by the official scoring protocol. We then compute the Mean Absolute Error (MAE) between the model-predicted and ground-truth scores as $\operatorname{MAE} = \frac{1}{N \times R} \sum_{i=1}^{N} \sum_{r=1}^{R} \lvert s_i(o_r) - \hat{s}_i(o_r) \rvert$, where $N$ is the number of items, $R$ is the number of response options, $s_i(o_r)$ is the ground-truth score, and $\hat{s}_i(o_r)$ is the model-predicted score for option $o_r$ of item $i$. A lower MAE indicates a better understanding of the inventory’s scoring protocol.

\subsection{Target Score Matching}
\paragraph{Evaluating LLMs' Target Score Matching in Psychometric Inventories.} We hypothesize that if LLMs have memorized the exact wording and the evaluation procedure of inventories, they would be able to strategically adjust responses to achieve desired scores. To test this, we prompt models with an item and a target score, instructing them to produce a response that would yield the specified score.
We assess target score matching using the Mean Absolute Error (MAE), defined as $\textstyle\operatorname{MAE} = \frac{1}{N}\sum_{i=1}^{N}|t_i - \hat{t}_i|$, where $N$ is the number of items in the inventory, $t_i$ is the target score for the $i$-th item, and $\hat{t}_i$ is the score achieved by the model. A lower MAE indicates a stronger capability of manipulating responses to match the desired target score. For each item, we conduct experiments with three target score conditions: minimum, mean, and maximum (e.g., 1, 3, and 5 for a five-point scale) and average the scores.

\section{Results\label{sec:results}}
\paragraph{Experimental Details.} 
We evaluate psychometric contamination on four widely used questionnaires: BFI-44, PVQ-40, MFQ, and SD-3, which measure personality traits, human values, moral foundations, and dark triad behaviors. These inventories are selected because they have been frequently adopted in recent NLP studies evaluating LLMs. We evaluate 21 models. The complete list of models, questionnaires, and prompt templates is provided in Appendix~\ref{appendix:experimental_details}. For all experiments we set the temperature to 0.7, repeat three times, and report the averaged results alongside 95 percent confidence intervals.

\begin{table*}[t]
\centering
{\small
\setlength{\tabcolsep}{3pt}
\renewcommand{\arraystretch}{1.0}
\begin{tabular}{@{}l c c c c c@{}}
\toprule
\textbf{Model} &
\multicolumn{2}{c}{\textbf{Item Memorization}} &
\multicolumn{2}{c}{\textbf{Evaluation Memorization}} &
\multicolumn{1}{c}{\textbf{Target Score Matching}} \\
\cmidrule(lr){2-3} \cmidrule(lr){4-5} \cmidrule(lr){6-6}
& \makecell{\textbf{Semantic}\\\textbf{Memorization}\\\textbf{(Similarity~$\uparrow$)}} &
  \makecell{\textbf{Key Information}\\\textbf{Memorization}\\\textbf{(Success Rate~$\uparrow$)}} &
  \makecell{\textbf{Item-Dimension}\\\textbf{Mapping}\\\textbf{(F1~$\uparrow$)}} &
  \makecell{\textbf{Option-Score}\\\textbf{Mapping}\\\textbf{(MAE~$\downarrow$)}} &
  \makecell{\textbf{(MAE~$\downarrow$)}} \\
\midrule
\multicolumn{6}{l}{\textit{OpenAI}}\\
gpt-4o-mini & 0.23 $\pm$ 0.01 & 0.33 $\pm$ 0.04 & 0.95 $\pm$ 0.03 & 0.73 $\pm$ 0.24 & 0.53 $\pm$ 0.04 \\
gpt-4o & 0.40 $\pm$ 0.02 & 0.40 $\pm$ 0.06 & 0.98 $\pm$ 0.01 & 0.10 $\pm$ 0.13 & 0.77 $\pm$ 0.17 \\
gpt-4.1-nano & 0.21 $\pm$ 0.01 & 0.27 $\pm$ 0.07 & 0.92 $\pm$ 0.07 & 0.85 $\pm$ 0.29 & 1.21 $\pm$ 0.08 \\
gpt-4.1-mini & 0.30 $\pm$ 0.01 & 0.33 $\pm$ 0.11 & 0.93 $\pm$ 0.02 & 0.21 $\pm$ 0.28 & 0.69 $\pm$ 0.11 \\
gpt-4.1 & 0.36 $\pm$ 0.02 & 0.46 $\pm$ 0.03 & 0.99 $\pm$ 0.00 & 0.32 $\pm$ 0.13 & 0.44 $\pm$ 0.12 \\
gpt-5-nano & 0.28 $\pm$ 0.05 & 0.34 $\pm$ 0.12 & 0.96 $\pm$ 0.03 & 0.12 $\pm$ 0.07 & 0.10 $\pm$ 0.06 \\
gpt-5-mini & 0.43 $\pm$ 0.02 & 0.54 $\pm$ 0.03 & 0.98 $\pm$ 0.03 & 0.05 $\pm$ 0.00 & 0.12 $\pm$ 0.03 \\
gpt-5 & 0.56 $\pm$ 0.02 & 0.79 $\pm$ 0.10 & 1.00 $\pm$ 0.01 & 0.01 $\pm$ 0.04 & 0.10 $\pm$ 0.02 \\
\midrule
\multicolumn{6}{l}{\textit{Qwen3}}\\
qwen3-14b & 0.20 $\pm$ 0.01 & 0.29 $\pm$ 0.09 & 0.94 $\pm$ 0.03 & 0.21 $\pm$ 0.11 & 0.15 $\pm$ 0.11 \\
qwen3-32b & 0.21 $\pm$ 0.01 & 0.30 $\pm$ 0.10 & 0.96 $\pm$ 0.05 & 0.28 $\pm$ 0.25 & 0.23 $\pm$ 0.12 \\
qwen3-235b-a22b & 0.34 $\pm$ 0.01 & 0.18 $\pm$ 0.03 & 0.78 $\pm$ 0.19 & 1.06 $\pm$ 0.69 & 0.12 $\pm$ 0.06 \\
\midrule
\multicolumn{6}{l}{\textit{GLM}}\\
glm-4-32b & 0.20 $\pm$ 0.01 & 0.26 $\pm$ 0.03 & 0.93 $\pm$ 0.01 & 0.41 $\pm$ 0.28 & 0.67 $\pm$ 0.10 \\
glm-4.5-air & 0.37 $\pm$ 0.02 & 0.31 $\pm$ 0.09 & 0.94 $\pm$ 0.04 & 0.16 $\pm$ 0.22 & 0.29 $\pm$ 0.34 \\
glm-4.5 & 0.41 $\pm$ 0.02 & 0.30 $\pm$ 0.13 & 0.97 $\pm$ 0.06 & 0.12 $\pm$ 0.06 & 0.13 $\pm$ 0.08 \\
\midrule
\multicolumn{6}{l}{\textit{Gemini}}\\
gemini-2.0-flash-001 & 0.32 $\pm$ 0.01 & 0.45 $\pm$ 0.08 & 0.82 $\pm$ 0.04 & 0.21 $\pm$ 0.05 & 0.44 $\pm$ 0.10 \\
gemini-2.5-flash-lite & 0.25 $\pm$ 0.01 & 0.33 $\pm$ 0.08 & 0.93 $\pm$ 0.03 & 0.32 $\pm$ 0.00 & 0.81 $\pm$ 0.07 \\
gemini-2.5-flash & 0.33 $\pm$ 0.01 & 0.41 $\pm$ 0.09 & 0.95 $\pm$ 0.05 & 0.18 $\pm$ 0.24 & 0.56 $\pm$ 0.10 \\
\midrule
\multicolumn{6}{l}{\textit{Claude}}\\
claude-3.5-sonnet & 0.34 $\pm$ 0.13 & 0.65 $\pm$ 0.10 & 0.99 $\pm$ 0.00 & 0.05 $\pm$ 0.03 & 0.35 $\pm$ 0.05 \\
claude-sonnet-4.5 & 0.46 $\pm$ 0.02 & 0.64 $\pm$ 0.03 & 0.98 $\pm$ 0.01 & 0.16 $\pm$ 0.01 & 0.47 $\pm$ 0.02 \\
\midrule
\multicolumn{6}{l}{\textit{Llama}}\\
llama-3.1-70b-instruct & 0.10 $\pm$ 0.02 & 0.35 $\pm$ 0.11 & 0.91 $\pm$ 0.05 & 1.34 $\pm$ 0.54 & 0.69 $\pm$ 0.13 \\
llama-3.1-405b-instruct & 0.16 $\pm$ 0.03 & 0.33 $\pm$ 0.10 & 0.88 $\pm$ 0.07 & 2.29 $\pm$ 0.45 & 0.60 $\pm$ 0.15 \\
\midrule
\textbf{Average} & 0.31 & 0.39 & 0.94 & 0.44 & 0.45 \\
\bottomrule
\end{tabular}
}
\caption{Psychometric contamination results averaged across BFI-44, MFQ, PVQ-40, and SD-3. Values are reported as mean $\pm$ 95\% confidence interval (CI) across three evaluation runs. Higher cosine similarity, success rate, and F1, and lower MAE indicate stronger contamination.}
\label{tab:psychometric-contamination}
\end{table*}

\paragraph{Overall Contamination Patterns.}
We begin with a high-level overview of contamination patterns across all models and inventories 
(Table~\ref{tab:psychometric-contamination}).  
Across the 21 LLMs evaluated, contamination is evident in all three aspects---\textit{item memorization}, 
\textit{evaluation memorization}, and \textit{target score matching}. Overall, models exhibit near-ceiling performance in \textit{evaluation memorization}, 
particularly in recognizing the psychological dimension each item measures. Moreover, models not only understand the evaluation procedures of inventories but also demonstrate the ability to strategically select response options to match specified target scores.

\paragraph{Item Memorization.}
The \textit{semantic memorization} task yields an average embedding similarity of 0.31. The \textit{key information memorization} reveals higher semantic retention (average success rate of 0.39), suggesting that models are aware of the core semantic content of psychometric items.

\paragraph{Evaluation Memorization.}
Understanding of the evaluation protocol appears to have saturated for most models, with \textit{item-dimension mapping} achieving an average F1-score of 0.94 across all 21 models. This indicates that LLMs already know which psychological construct each item measures. In \textit{option–score mapping}, higher-performing models (e.g., GPT-5, GLM-4.5) exhibit low MAE, showing that they have internalized scoring schemes, including reverse-coded items.

\paragraph{Target Score Matching.}
Advanced models demonstrate the ability to achieve desired target scores 
(MAE $\approx$ 0.1--0.2). This suggests that memorization can be extended beyond simple recall to the application of learned scoring logic.

\paragraph{Average Trends and Family Behaviors.}
On average, contamination appears to be saturated for recognizing item-dimension mapping and scoring knowledge. Recent model families such as GPT-5, Claude, and GLM exhibit the most consistent contamination pattern, while Llama variants show less. As shown in Table~\ref{tab:questionnaire-wise}, PVQ-40 and BFI-44 display stronger contamination compared to MFQ and SD-3, likely reflecting their greater online availability and more frequent use in NLP research. For more detailed analysis, see Appendix~\ref{appendix:additional_results}.

\paragraph{Scaling Effects.}
Larger models within the same family exhibit lower MAE and higher F1-scores, suggesting that scaling may amplify contamination. However, this effect diminishes once models reach saturation in evaluation memorization. In contrast, smaller models show comparatively weaker contamination, suggesting that psychometric familiarity emerges progressively with increasing model size and training exposure.

\paragraph{Implications.}
Across 21 models and four widely used psychometric inventories, our results provide systematic evidence that LLMs not only recall inventory items but also understand their scoring procedures and can strategically generate responses to achieve specific target scores. These findings offer the first systematic experimental evidence supporting prior concerns that psychometric evaluations of LLMs may suffer from data contamination. Moreover, since contamination is more pronounced in widely used inventories such as BFI-44 and PVQ-40, our results underscore the need for contamination-aware evaluation practices.

\paragraph{Recommendations.\label{para:recommendations}}
Because LLMs may have been exposed to many psychometric inventories available online, directly comparing the magnitude of contamination signals across settings is not straightforward. To facilitate interpretation, we define simple baselines for each contamination setting and provide practical guidelines for interpreting the results.

For \textit{semantic memorization}, we use \textit{dimension description text} as a baseline (i.e., a short definition of the trait/value that the item is measuring) and measure its embedding similarity to the item text. This baseline yields cosine similarities of 0.32 (BFI-44), 0.30 (MFQ), 0.41 (PVQ-40), and 0.19 (SD3), with an overall average of 0.32 across inventories. If a model's semantic memorization score is close to the baseline, it suggests that the model's reconstruction is at the level of generic dimension definition.

For \textit{key information memorization}, establishing a baseline corresponding to a truly uncontaminated model is inherently challenging, since there is no well-defined procedure for approximating random keyword generation under conditions of potential contamination. We therefore contextualize the results in Table~\ref{tab:psychometric-contamination} using the empirical distribution of performance across the 21 evaluated models. Specifically, under the assumption that lower-scoring models are likely to exhibit less contamination, we report the 10th-percentile success rates: 0.16 (BFI-44), 0.25 (MFQ), 0.41 (PVQ-40), and 0.18 (SD-3). These values should be interpreted cautiously, however, as even low-scoring models may exhibit some degree of contamination. We adopt the 10th percentile to reduce sensitivity to outliers, but this choice is inherently heuristic, and alternative percentile thresholds could be adopted depending on the practitioner’s objectives.

Regarding absolute values, the success rate has a direct operational meaning where 0 indicates that the model never recovers the important keyword of an item, whereas 1 indicates that it recovers it for all items. Intermediate values can be interpreted as frequency. For example, a score of 0.5 means that the model knows the exact wording of the important keyword for about half of the items, suggesting that the model knows the semantics of at least half of the items in an inventory.

For \textit{item-dimension mapping}, a random classifier over $N$ dimensions yields an expected F1 of $1/N$ (0.2 for BFI-44 with $N=5$). An F1 score near the baseline suggests the model is simply guessing.

For \textit{option-score mapping}, if a model assigns scores uniformly at random on a $K$-point scale inventory, the expected mean absolute error is $(K^2-1)/(3K)$ (1.60 for $K=5$ and 1.94 for $K=6$). MAE score near this baseline indicates the model is guessing the score associated with each response option.

For \textit{target score matching}, we use a random baseline where the model selects a response uniformly at random from a $K$-point scale inventory, independent of the target. Under this strategy, the expected MAE is $\mathbb{E}[|t-\hat t|]=\frac{(t-1)t+(K-t)(K-t+1)}{2K}$, where $t$ is the target score and $\hat t$ is the randomly selected score. Averaging over the three target conditions (minimum, mean, and maximum) yields expected MAE values of 1.73 for $K=5$ and 2.17 for $K=6$. An MAE close to the baseline indicates that the model is guessing randomly.

\section{Conclusion\label{sec:conclusion}}
In this work, we propose a framework to quantify data contamination in psychometric evaluations of LLMs. Our analysis covers three aspects: (1) \textit{item memorization}, (2) \textit{evaluation memorization}, and (3) \textit{target score matching}---covering 21 models and four widely used inventories. Our results provide the first systematic evidence that contemporary LLMs already possess substantial psychometric knowledge, accurately recalling questionnaire items, understanding scoring schemes, and even manipulating responses to achieve desired scores.

\section*{Limitations\label{sec:limitations}}
While our study analyzes data contamination across several questionnaires commonly used in NLP, many other psychometric inventories remain unexplored. In addition, our analysis is limited to English items and may not generalize to other languages or culturally adapted versions. Future work could extend our framework to examine how data contamination affects evaluation outcomes and to explore other related phenomena that may correlate with contamination in LLMs.

\section{Acknowledgments}
This work was supported by the Creative-Pioneering Researchers Program through Seoul National University and by the National Research Foundation of Korea (NRF) grant funded by the Korean government (MSIT) (RS-2024-00333484). It was also supported by the National Supercomputing Center with supercomputing resources including technical support (KSC-2025-CRE-0332, KSC-2025-CRE-0514).

\bibliography{custom}

\appendix

\section{Related Work\label{appendix:related_work}}

\subsection{Psychometric Inventories}
\paragraph{Big Five Inventory.} The Big Five Inventory (BFI) measures personality across five dimensions: Openness, Conscientiousness, Extraversion, Agreeableness, and Neuroticism~\citep{BFI-10}. It is one of the most extensively studied frameworks in psychology and has been widely used to assess individual personality traits. Recently, the BFI has also been applied to large language models (LLMs) to characterize their behavioral tendencies, evaluate their alignment with human-like personality profiles, and explore their potential applications in agent-based simulations~\citep{jiang-etal-2024-personallm}.

\paragraph{Portrait Values Questionnaire.}
Schwartz's theory of basic human values defines ten value dimensions that are considered universal across cultures~\citep{schwartz2012overview}. These dimensions are: Achievement, Benevolence, Conformity, Security, Stimulation, Self-Direction, Tradition, Hedonism, Universalism, and Power. The Portrait Values Questionnaire (PVQ) operationalizes this framework by presenting short verbal portraits describing people’s goals and aspirations, which respondents evaluate based on similarity to themselves. PVQ has recently been adopted in LLM studies to examine how models express or prioritize human-like value orientations~\citep{choi2025establishedpsychometricvsecologically}.

\paragraph{Moral Foundations Questionnaire.}
The Moral Foundations Questionnaire (MFQ) is based on moral foundations theory, which proposes five core dimensions of human morality: Harm/Care, Fairness/Reciprocity, In-group/Loyalty, Authority/Respect, and Purity/Sanctity~\citep{MFQ-30}. The MFQ consists of two sections that measure moral relevance and moral judgment, and includes two attention-check items (the 6th and 22nd) designed to identify respondents who answer carelessly or randomly. This questionnaire has been recently used to examine the moral reasoning patterns and normative biases of LLMs~\citep{abdulhai-etal-2024-moral}.

\paragraph{Short Dark Triad.}
The Short Dark Triad (SD-3) captures three socially aversive personality traits collectively known as the ``dark triad'': Machiavellianism, Narcissism, and Psychopathy~\citep{sd3}. It serves as a concise alternative to longer instruments such as the Mach-IV, NPI, and SRP scales, while maintaining strong psychometric validity. Recently, the SD-3 has been directly applied or adapted for use with LLMs to examine whether models exhibit or simulate behaviors associated with manipulativeness, self-centeredness, and reduced empathy~\citep{li-etal-2024-evaluating-psychological, lee-etal-2025-llms}.

\subsection{Applying Psychometric Inventories in LLM Evaluation}
The mass adoption of LLMs has motivated growing interest in evaluating their alignment with human cognition and behavior~\citep{lim2025psychometricitemvalidationusing}. A common approach in this line of work is to assess LLMs using established psychometric inventories and compare their results with those of humans.

\citet{miotto-etal-2022-gpt} applied the HEXACO personality scale and Human Values Scale to GPT-3, showing that the model exhibits personality profiles similar to human respondents. \citet{Hadar_Shoval} used the Big Five Inventory (BFI), while \citet{tlaie2024exploring} employed the Moral Foundations Questionnaire (MFQ) to examine moral and personality-related response patterns in LLMs. Similarly, \citet{info15110679} used the Portrait Values Questionnaire (PVQ) to analyze value preferences in LLMs.

\subsection{Data Contamination and Its Implications for Benchmarking LLMs\label{sec:data_contamination}}
Data contamination poses a significant risk to the benchmarking and evaluation of LLMs, undermining the reliability of performance assessments. Because LLMs are trained on massive text corpora, there is a possibility that evaluation datasets overlap with training data, leading to unreliable evaluation results~\cite{balloccu-etal-2024-leak}. This issue is particularly problematic when models exhibit strong results on benchmarks, as it becomes unclear whether these outcomes stem from genuine generalization or mere memorization of test data \cite{sainz-etal-2023-nlp}. In the field of computational psychology, contamination is especially concerning because psychometric inventories are easily accessible in academic papers and online repositories and may have been included in pretraining data.

To address this challenge, several studies have proposed methods to detect and mitigate data contamination. \citet{dong-etal-2024-generalization} proposed CDD (Contamination Detection via Output Distribution), and TED (Trustworthy Evaluation via Output Distribution), both using output distribution patterns to identify and correct for contamination. \citet{fu-etal-2025-data} surveyed detection techniques such as verbatim memorization and key information generation. In this paper, we build on these contamination detection approaches to investigate whether LLMs have internalized psychometric inventories, thereby questioning the suitability of such tools for measuring value orientation.

\section{Experimental Details\label{appendix:experimental_details}}

\subsection{Models}
We evaluate models from the following major families: GPT-4 (4o-mini, 4o, 4.1-nano, 4.1-mini, 4.1), GPT-5 (nano, mini, base), Qwen3 (32B, 235B), GLM (4-32B, 4.5 Air, 4.5), Gemini (2.0 Flash, 2.5 Flash Lite, 2.5 Flash), Claude (3.5 Sonnet, 4.5 Sonnet), and Llama 3.1 (70B, 405B). All models are evaluated in their instruction-tuned variants when available, with the temperature set to 0.7. All experiments are conducted via the OpenRouter API, with a total cost of approximately 50 USD.

\subsection{Prompts}

\begin{lstlisting}[caption={Prompt 1: Prompt template for semantic memorization.},  label={lst: semantic}]
You are given metadata about a psychometric inventory item.
Your task is to reproduce the exact full wording of that item.

Metadata:
- Inventory name: {inventory_name}
- Item index: {item_index}  (Note: indexing starts from 1)

Instructions:
- Output ONLY the exact item text, with no numbering, labels, or extra words.
- Preserve the original wording, casing, and punctuation if you know it.
\end{lstlisting}
\begin{lstlisting}[caption={Prompt 2: Prompt template for refusal detection and similarity evaluation.},  label={lst: refusal-sim}]
You are given two texts: (1) the original text and (2) the candidate text.
Your task is to evaluate the candidate text according to two criteria.

Task 1 -- Refusal Detection
Determine whether the candidate text is a refusal.
A refusal means the model declines, deflects, or avoids performing the task (e.g., I cannot answer, I'm unable to help, I don't have enough information).
Respond with Yes or No.

Task 2 -- Similarity Evaluation
Evaluate how similar the candidate text is to the original text.
Use the following options exactly as written:

Highly Similar, Moderately Similar, Slightly Similar, Slightly Dissimilar, Moderately Dissimilar, Highly Dissimilar

Similarity should consider meaning, intent, and content---not surface-level phrasing.

Original text: {original}
Candidate text: {candidate}
\end{lstlisting}

\begin{lstlisting}[caption={Prompt 3: Prompt template for keyitem generation.},  label={lst: keyitem}]
You are given a psychometric inventory item with one important keyword masked as [MASK].
Your task is to guess the masked keyword.

Item (masked):
{item_with_mask}

Instructions:
- Output ONLY the single masked keyword.
- Do NOT include quotes, punctuation, or any extra text.
\end{lstlisting}

\begin{lstlisting}[caption={Prompt 4: Prompt template for item-dimension mapping.},  label={lst: item-dimension}]
You are given a psychometric inventory, and the dimensions it measures.
Your task is to map a given item to the correct dimension.

Inventory name: {inventory_name}
Dimensions: {dimensions}

Item:
{item}

Instructions:
- For the item, return ONLY the corresponding dimension name.
- Use exactly the dimension labels provided in the metadata.
- Do NOT include quotes, punctuation, or any extra text.
\end{lstlisting}

\begin{lstlisting}[caption={Prompt 5: Prompt template for option-score mapping.},  label={lst: option-score}]
You are given a psychometric item, its target dimension, and the response options.
Your task is to map each option to the numeric score it contributes to the target dimension.

Item: {item}
Dimension: {dimension}
Options: {options}


Instructions:
- Output only the numeric scores for the options in order, separated by commas.
- Apply reverse-coding if the item requires it.
- All scores are positive integers.
- Do NOT include quotes, punctuation, or any extra text.

Output format:
score_1, score_2, score_3, ...
\end{lstlisting}

\begin{lstlisting}[caption={Prompt 6: Prompt template for target score mapping.},  label={lst: target score}]
You are given a psychometric item, its target dimension, and a target score.
Your task is to select the response option that corresponds to the given target score.

Item: {item}
Dimension: {dimension}
Target Score: {target_score}
Options: {options}

Instructions:
- Choose the option that produces the target score.
- Apply reverse-coding if necessary.
- Output only the chosen option, with no extra text.
\end{lstlisting}

\section{Additional Results and Analyses\label{appendix:additional_results}}
Table~\ref{tab:questionnaire-wise} presents the comparison of data contamination across questionnaires. Tables~\ref{tab:bfi}, \ref{tab:mfq}, \ref{tab:pvq}, and~\ref{tab:sd3} show the results on the BFI-44, MFQ, PVQ-40, and SD-3 inventories.

As mentioned in Section~\ref{sec:item_memorization}, we filtered out refusal responses with GPT-5.2-mini. We used Prompt~\ref{lst: refusal-sim} to detect refusal responses. In the same prompt, we also asked the model to grade the similarity between the reference text and the generated text on a 6-point scale, ranging from highly dissimilar to highly similar. Tables~\ref{tab:bfi44_refusal_ci}, \ref{tab:mfq_refusal_ci}, \ref{tab:pvq40_refusal_ci}, and \ref{tab:sd3_refusal_ci} show the refusal rates and similarity scores on the BFI-44, MFQ, PVQ-40, and SD-3 inventories.

\paragraph{Contamination differences across questionnaires.}
In Section~\ref{sec:results}, we observed that PVQ-40 and BFI-44 exhibit stronger contamination signals than MFQ and SD-3, and hypothesized that this difference may be related to their greater online and academic exposure. To further justify this interpretation, we examined the citation counts of the primary papers associated with each inventory. The results are as follows: BFI (12,560)~\cite{BFI-10}, PVQ (7,087)~\cite{schwartz2012overview}, MFQ (4,226)~\cite{MFQ-30}, and SD3 (3,369)~\cite{sd3}, as of 2025. While citation counts are an approximate indicator of exposure, these differences suggest that BFI and PVQ have broader academic visibility than MFQ and SD-3, which may contribute to greater exposure during model training.

\paragraph{Baseline settings.}
For the semantic memorization baseline, we use the dimension description text for each inventory. For BFI-44, we use the trait descriptions from \citet{pervin1999handbook}. For MFQ, we use the foundation descriptions provided on \url{https://moralfoundations.org/}. For PVQ-40, we use the value descriptions from \citet{schwartz2012overview}. For SD-3, we use dimension descriptions adapted from \citet{sd3}.

\begin{lstlisting}[caption={Dimension definitions of inventories used in semantic memorization baseline experiments.},  label={lst: definitions}]
#BFI-44
Openness: Ideas (curious), fantasy (imaginative), aesthetics (artistic), actions (wide interests), feelings (excitable), and values (unconventional).

Conscientiousness: Competence (efficient), order (organized), dutifulness (not careless), achievement striving (thorough), self-discipline (not lazy), and deliberation (not impulsive).

Extraversion: Gregariousness (sociable), assertiveness (forceful), activity (energetic), excitement-seeking (adventurous), positive emotions (enthusiastic), and warmth (outgoing).

Agreeableness: Trust (forgiving), straightforwardness (not demanding), altruism (warm), compliance (not stubborn), modesty (not show-off), and tender-mindedness (sympathetic).

Neuroticism: Anxiety (tense), angry hostility (irritable), depression (not contented), self-consciousness (shy), impulsiveness (moody), and vulnerability (not self-confident).

#MFQ
Care: This foundation is related to our long evolution as mammals with attachment systems and an ability to feel (and dislike) the pain of others. It underlies the virtues of kindness, gentleness, and nurturance.

Fairness: This foundation is related to the evolutionary process of reciprocal altruism. It underlies the virtues of justice and rights.

Loyalty: This foundation is related to our long history as tribal creatures able to form shifting coalitions. It is active anytime people feel that it's one for all and all for one. It underlies the virtues of patriotism and self-sacrifice for the group.

Authority: This foundation was shaped by our long primate history of hierarchical social interactions. It underlies virtues of leadership and followership, including deference to prestigious authority figures and respect for traditions.

Purity: This foundation was shaped by the psychology of disgust and contamination. It underlies notions of striving to live in an elevated, less carnal, more noble, and more natural way (often present in religious narratives). This foundation underlies the widespread idea that the body is a temple that can be desecrated by immoral activities and contaminants (an idea not unique to religious traditions). It underlies the virtues of self-discipline, self-improvement, naturalness, and spirituality.

#PVQ-40
Self-Direction: Independent thought and action--choosing, creating, and exploring.

Stimulation: Excitement, novelty, and challenge in life.

Hedonism: Pleasure or sensuous gratification for oneself.

Achievement: Personal success through demonstrating competence according to social standards.

Power: Social status and prestige, control or dominance over people and resources.

Security: Safety, harmony, and stability of society, of relationships, and of self.

Conformity: Restraint of actions, inclinations, and impulses likely to upset or harm others and violate social expectations or norms.

Tradition: Respect, commitment, and acceptance of the customs and ideas that one's culture or religion provides.

Benevolence: Preserving and enhancing the welfare of those with whom one is in frequent personal contact (the 'in-group').

Universalism: Understanding, appreciation, tolerance, and protection for the welfare of all people and for nature.

#SD-3
Machiavellianism: Manipulativeness, callous affect, and a strategic-calculating orientation.

Psychopathy: Deficits in affect (i.e., callousness) and self-control (i.e., impulsivity).

Narcissism: Ego-identity goals drive narcissistic behavior.
\end{lstlisting}

\begin{table*}[htbp]
\centering
{\small
\setlength{\tabcolsep}{5pt}
\renewcommand{\arraystretch}{1.0}
\begin{tabular}{@{}l c c c c c@{}}
\toprule
\textbf{Questionnaire} &
\makecell{\textbf{Semantic}\\\textbf{Memorization}\\\textbf{(Similarity~$\uparrow$)}} &
\makecell{\textbf{Key Information}\\\textbf{Memorization}\\ \textbf{(Success Rate~$\uparrow$)}} &
\makecell{\textbf{Item-Dimension}\\\textbf{Mapping}\\\textbf{(F1~$\uparrow$)}} &
\makecell{\textbf{Option-Score}\\\textbf{Mapping}\\\textbf{(MAE~$\downarrow$)}} &
\makecell{\textbf{Target Score}\\\textbf{Matching}\\\textbf{(MAE~$\downarrow$)}} \\
\midrule
BFI-44 & 0.36 & 0.34 & \textbf{0.96} & 0.41 & 0.37 \\
MFQ & 0.17 & 0.35 & 0.95 & 0.70 & 0.68 \\
PVQ-40 & \textbf{0.37} & \textbf{0.57} & 0.91 & \textbf{0.28} & \textbf{0.35} \\
SD-3 & 0.24 & 0.31 & 0.93 & 0.37 & 0.41 \\
\bottomrule
\end{tabular}
}
\caption{Questionnaire-wise contamination comparison (model averages). Lower AED and MAE and higher success rate and F1-score reflect greater contamination.}
\label{tab:questionnaire-wise}
\end{table*}

\begin{table*}[t]
\centering
{\small
\setlength{\tabcolsep}{5pt}
\renewcommand{\arraystretch}{1.0}
\begin{tabular}{@{}l c c c c c@{}}
\toprule
\textbf{Model} &
\multicolumn{2}{c}{\textbf{Item Memorization}} &
\multicolumn{2}{c}{\textbf{Evaluation Memorization}} &
\multicolumn{1}{c}{\textbf{Target Score Matching}} \\
\cmidrule(lr){2-3} \cmidrule(lr){4-5} \cmidrule(lr){6-6}
& \makecell{\textbf{Semantic}\\\textbf{Memorization}\\\textbf{(Similarity~$\uparrow$)}} &
  \makecell{\textbf{Key Information}\\\textbf{Memorization}\\\textbf{(Success Rate~$\uparrow$)}} &
  \makecell{\textbf{Item--Dimension}\\\textbf{Mapping}\\\textbf{(F1~$\uparrow$)}} &
  \makecell{\textbf{Option--Score}\\\textbf{Mapping}\\\textbf{(MAE~$\downarrow$)}} &
  \makecell{\textbf{(MAE~$\downarrow$)}} \\
\midrule
\multicolumn{6}{l}{\textit{OpenAI}}\\
gpt-4o-mini & 0.35 $\pm$ 0.02 & 0.17 $\pm$ 0.09 & 0.96 $\pm$ 0.00 & 0.25 $\pm$ 0.08 & 0.56 $\pm$ 0.04 \\
gpt-4o & 0.44 $\pm$ 0.04 & 0.36 $\pm$ 0.06 & 1.00 $\pm$ 0.00 & 0.00 $\pm$ 0.00 & 0.38 $\pm$ 0.11 \\
gpt-4.1-nano & 0.37 $\pm$ 0.02 & 0.16 $\pm$ 0.06 & 0.92 $\pm$ 0.06 & 1.61 $\pm$ 0.29 & 0.73 $\pm$ 0.12 \\
gpt-4.1-mini & 0.21 $\pm$ 0.02 & 0.20 $\pm$ 0.15 & 0.98 $\pm$ 0.00 & 0.02 $\pm$ 0.10 & 0.69 $\pm$ 0.18 \\
gpt-4.1 & 0.25 $\pm$ 0.03 & 0.41 $\pm$ 0.00 & 1.00 $\pm$ 0.00 & 0.83 $\pm$ 0.21 & 0.36 $\pm$ 0.14 \\
gpt-5-nano & 0.40 $\pm$ 0.07 & 0.25 $\pm$ 0.11 & 0.98 $\pm$ 0.02 & 0.00 $\pm$ 0.00 & 0.00 $\pm$ 0.01 \\
gpt-5-mini & 0.58 $\pm$ 0.04 & 0.57 $\pm$ 0.00 & 1.00 $\pm$ 0.00 & 0.00 $\pm$ 0.00 & 0.00 $\pm$ 0.00 \\
gpt-5 & 0.67 $\pm$ 0.02 & 0.84 $\pm$ 0.11 & 1.00 $\pm$ 0.00 & 0.00 $\pm$ 0.00 & 0.00 $\pm$ 0.00 \\
\midrule
\multicolumn{6}{l}{\textit{Qwen3}}\\
qwen3-14b & 0.28 $\pm$ 0.02 & 0.21 $\pm$ 0.03 & 0.98 $\pm$ 0.00 & 0.00 $\pm$ 0.00 & 0.03 $\pm$ 0.08 \\
qwen3-32b & 0.29 $\pm$ 0.01 & 0.18 $\pm$ 0.11 & 0.97 $\pm$ 0.04 & 0.00 $\pm$ 0.00 & 0.04 $\pm$ 0.09 \\
qwen3-235b-a22b & 0.28 $\pm$ 0.03 & 0.14 $\pm$ 0.06 & 0.81 $\pm$ 0.16 & 0.86 $\pm$ 0.59 & 0.00 $\pm$ 0.00 \\
\midrule
\multicolumn{6}{l}{\textit{GLM}}\\
glm-4-32b & 0.27 $\pm$ 0.02 & 0.12 $\pm$ 0.03 & 0.93 $\pm$ 0.00 & 0.15 $\pm$ 0.28 & 0.81 $\pm$ 0.06 \\
glm-4.5-air & 0.47 $\pm$ 0.03 & 0.23 $\pm$ 0.17 & 0.98 $\pm$ 0.00 & 0.08 $\pm$ 0.33 & 0.21 $\pm$ 0.82 \\
glm-4.5 & 0.57 $\pm$ 0.05 & 0.32 $\pm$ 0.06 & 0.99 $\pm$ 0.03 & 0.00 $\pm$ 0.00 & 0.00 $\pm$ 0.00 \\
\midrule
\multicolumn{6}{l}{\textit{Gemini}}\\
gemini-2.0-flash-001 & 0.27 $\pm$ 0.02 & 0.52 $\pm$ 0.06 & 0.98 $\pm$ 0.00 & 0.16 $\pm$ 0.14 & 0.41 $\pm$ 0.07 \\
gemini-2.5-flash-lite & 0.26 $\pm$ 0.02 & 0.21 $\pm$ 0.03 & 0.98 $\pm$ 0.00 & 0.87 $\pm$ 0.00 & 1.04 $\pm$ 0.08 \\
gemini-2.5-flash & 0.22 $\pm$ 0.02 & 0.39 $\pm$ 0.11 & 0.98 $\pm$ 0.06 & 0.05 $\pm$ 0.00 & 0.36 $\pm$ 0.07 \\
\midrule
\multicolumn{6}{l}{\textit{Claude}}\\
claude-3.5-sonnet & 0.29 $\pm$ 0.37 & 0.73 $\pm$ 0.11 & 1.00 $\pm$ 0.02 & 0.00 $\pm$ 0.00 & 0.28 $\pm$ 0.08 \\
claude-sonnet-4.5 & 0.53 $\pm$ 0.09 & 0.67 $\pm$ 0.03 & 1.00 $\pm$ 0.00 & 0.00 $\pm$ 0.00 & 0.46 $\pm$ 0.04 \\
\midrule
\multicolumn{6}{l}{\textit{Llama}}\\
llama-3.1-70b-instruct & 0.21 $\pm$ 0.08 & 0.19 $\pm$ 0.03 & 0.96 $\pm$ 0.03 & 1.40 $\pm$ 0.79 & 0.87 $\pm$ 0.22 \\
llama-3.1-405b-instruct & 0.20 $\pm$ 0.30 & 0.23 $\pm$ 0.03 & 0.87 $\pm$ 0.07 & 2.25 $\pm$ 0.17 & 0.53 $\pm$ 0.06 \\
\midrule
\textbf{Average} & 0.36 & 0.34 & 0.96 & 0.41 & 0.37 \\
\bottomrule
\end{tabular}
}
\caption{Psychometric contamination results for BFI-44 with 95\% confidence intervals from three experimental repetitions.}
\label{tab:bfi}
\end{table*}

\begin{table*}[t]
\centering
{\small
\setlength{\tabcolsep}{3pt}
\renewcommand{\arraystretch}{1.0}
\begin{tabular}{@{}l c c c c c@{}}
\toprule
\textbf{Model} &
\multicolumn{2}{c}{\textbf{Item Memorization}} &
\multicolumn{2}{c}{\textbf{Evaluation Memorization}} &
\multicolumn{1}{c}{\textbf{Target Score Matching}} \\
\cmidrule(lr){2-3} \cmidrule(lr){4-5} \cmidrule(lr){6-6}
& \makecell{\textbf{Semantic}\\\textbf{Memorization}\\\textbf{(Similarity~$\uparrow$)}} &
  \makecell{\textbf{Key Information}\\\textbf{Memorization}\\\textbf{(Success Rate~$\uparrow$)}} &
  \makecell{\textbf{Item--Dimension}\\\textbf{Mapping}\\\textbf{(F1~$\uparrow$)}} &
  \makecell{\textbf{Option--Score}\\\textbf{Mapping}\\\textbf{(MAE~$\downarrow$)}} &
  \makecell{\textbf{(MAE~$\downarrow$)}} \\
\midrule
\multicolumn{6}{l}{\textit{OpenAI}}\\
gpt-4o-mini & 0.11 $\pm$ 0.01 & 0.36 $\pm$ 0.04 & 0.93 $\pm$ 0.00 & 2.17 $\pm$ 0.52 & 0.57 $\pm$ 0.03 \\
gpt-4o & 0.37 $\pm$ 0.00 & 0.32 $\pm$ 0.04 & 0.97 $\pm$ 0.00 & 0.17 $\pm$ 0.14 & 1.12 $\pm$ 0.19 \\
gpt-4.1-nano & 0.10 $\pm$ 0.01 & 0.28 $\pm$ 0.08 & 0.90 $\pm$ 0.08 & 0.32 $\pm$ 0.10 & 1.54 $\pm$ 0.10 \\
gpt-4.1-mini & 0.10 $\pm$ 0.01 & 0.30 $\pm$ 0.09 & 0.95 $\pm$ 0.05 & 0.47 $\pm$ 0.38 & 0.90 $\pm$ 0.09 \\
gpt-4.1 & 0.32 $\pm$ 0.05 & 0.41 $\pm$ 0.00 & 1.00 $\pm$ 0.00 & 0.10 $\pm$ 0.00 & 0.43 $\pm$ 0.07 \\
gpt-5-nano & 0.12 $\pm$ 0.03 & 0.32 $\pm$ 0.09 & 0.99 $\pm$ 0.05 & 0.47 $\pm$ 0.14 & 0.31 $\pm$ 0.12 \\
gpt-5-mini & 0.22 $\pm$ 0.04 & 0.44 $\pm$ 0.00 & 1.00 $\pm$ 0.00 & 0.20 $\pm$ 0.00 & 0.43 $\pm$ 0.07 \\
gpt-5 & 0.31 $\pm$ 0.07 & 0.81 $\pm$ 0.21 & 1.00 $\pm$ 0.00 & 0.03 $\pm$ 0.14 & 0.39 $\pm$ 0.04 \\
\midrule
\multicolumn{6}{l}{\textit{Qwen3}}\\
qwen3-14b & 0.11 $\pm$ 0.01 & 0.25 $\pm$ 0.08 & 0.98 $\pm$ 0.05 & 0.77 $\pm$ 0.29 & 0.39 $\pm$ 0.15 \\
qwen3-32b & 0.10 $\pm$ 0.01 & 0.24 $\pm$ 0.09 & 0.98 $\pm$ 0.05 & 0.73 $\pm$ 0.38 & 0.76 $\pm$ 0.22 \\
qwen3-235b-a22b & 0.26 $\pm$ 0.03 & 0.14 $\pm$ 0.04 & 0.83 $\pm$ 0.19 & 1.41 $\pm$ 0.99 & 0.47 $\pm$ 0.19 \\
\midrule
\multicolumn{6}{l}{\textit{GLM}}\\
glm-4-32b & 0.11 $\pm$ 0.01 & 0.26 $\pm$ 0.04 & 0.93 $\pm$ 0.00 & 1.08 $\pm$ 0.27 & 0.72 $\pm$ 0.21 \\
glm-4.5-air & 0.12 $\pm$ 0.02 & 0.29 $\pm$ 0.04 & 0.98 $\pm$ 0.05 & 0.50 $\pm$ 0.25 & 0.74 $\pm$ 0.41 \\
glm-4.5 & 0.21 $\pm$ 0.04 & 0.27 $\pm$ 0.16 & 0.98 $\pm$ 0.07 & 0.44 $\pm$ 0.13 & 0.53 $\pm$ 0.32 \\
\midrule
\multicolumn{6}{l}{\textit{Gemini}}\\
gemini-2.0-flash-001 & 0.13 $\pm$ 0.02 & 0.39 $\pm$ 0.12 & 0.84 $\pm$ 0.00 & 0.48 $\pm$ 0.05 & 0.54 $\pm$ 0.08 \\
gemini-2.5-flash-lite & 0.10 $\pm$ 0.02 & 0.35 $\pm$ 0.12 & 0.87 $\pm$ 0.06 & 0.00 $\pm$ 0.00 & 0.91 $\pm$ 0.02 \\
gemini-2.5-flash & 0.26 $\pm$ 0.03 & 0.30 $\pm$ 0.04 & 0.94 $\pm$ 0.05 & 0.49 $\pm$ 0.82 & 0.79 $\pm$ 0.14 \\
\midrule
\multicolumn{6}{l}{\textit{Claude}}\\
claude-3.5-sonnet & 0.15 $\pm$ 0.66 & 0.52 $\pm$ 0.09 & 0.97 $\pm$ 0.00 & 0.21 $\pm$ 0.13 & 0.64 $\pm$ 0.03 \\
claude-sonnet-4.5 & 0.95 $\pm$ 0.00 & 0.46 $\pm$ 0.04 & 1.00 $\pm$ 0.00 & 0.66 $\pm$ 0.05 & 0.41 $\pm$ 0.03 \\
\midrule
\multicolumn{6}{l}{\textit{Llama}}\\
llama-3.1-70b-instruct & 0.09 $\pm$ 0.03 & 0.30 $\pm$ 0.16 & 0.92 $\pm$ 0.06 & 1.54 $\pm$ 0.76 & 0.76 $\pm$ 0.10 \\
llama-3.1-405b-instruct & 0.05 $\pm$ 0.19 & 0.35 $\pm$ 0.16 & 0.94 $\pm$ 0.03 & 2.39 $\pm$ 0.31 & 0.93 $\pm$ 0.31 \\
\midrule
\textbf{Average} & 0.17 & 0.35 & 0.95 & 0.70 & 0.68 \\
\bottomrule
\end{tabular}
}
\caption{Psychometric contamination results for MFQ with 95\% confidence intervals from three experimental repetitions.}
\label{tab:mfq}
\end{table*}

\begin{table*}[t]
\centering
{\small
\setlength{\tabcolsep}{3pt}
\renewcommand{\arraystretch}{1.0}
\begin{tabular}{@{}l c c c c c@{}}
\toprule
\textbf{Model} &
\multicolumn{2}{c}{\textbf{Item Memorization}} &
\multicolumn{2}{c}{\textbf{Evaluation Memorization}} &
\multicolumn{1}{c}{\textbf{Target Score Matching}} \\
\cmidrule(lr){2-3} \cmidrule(lr){4-5} \cmidrule(lr){6-6}
& \makecell{\textbf{Semantic}\\\textbf{Memorization}\\\textbf{(Similarity~$\uparrow$)}} &
  \makecell{\textbf{Key Information}\\\textbf{Memorization}\\\textbf{(Success Rate~$\uparrow$)}} &
  \makecell{\textbf{Item--Dimension}\\\textbf{Mapping}\\\textbf{(F1~$\uparrow$)}} &
  \makecell{\textbf{Option--Score}\\\textbf{Mapping}\\\textbf{(MAE~$\downarrow$)}} &
  \makecell{\textbf{(MAE~$\downarrow$)}} \\
\midrule
\multicolumn{6}{l}{\textit{OpenAI}}\\
gpt-4o-mini & 0.21 $\pm$ 0.01 & 0.54 $\pm$ 0.04 & 0.97 $\pm$ 0.04 & 0.25 $\pm$ 0.22 & 0.40 $\pm$ 0.03 \\
gpt-4o & 0.42 $\pm$ 0.03 & 0.58 $\pm$ 0.04 & 0.96 $\pm$ 0.04 & 0.07 $\pm$ 0.00 & 0.86 $\pm$ 0.06 \\
gpt-4.1-nano & 0.18 $\pm$ 0.01 & 0.01 $\pm$ 0.06 & 0.90 $\pm$ 0.06 & 0.80 $\pm$ 0.39 & 1.12 $\pm$ 0.03 \\
gpt-4.1-mini & 0.44 $\pm$ 0.02 & 0.54 $\pm$ 0.09 & 0.92 $\pm$ 0.02 & 0.10 $\pm$ 0.11 & 0.39 $\pm$ 0.00 \\
gpt-4.1 & 0.47 $\pm$ 0.02 & 0.62 $\pm$ 0.06 & 0.97 $\pm$ 0.00 & 0.03 $\pm$ 0.13 & 0.53 $\pm$ 0.16 \\
gpt-5-nano & 0.31 $\pm$ 0.11 & 0.45 $\pm$ 0.16 & 0.89 $\pm$ 0.03 & 0.03 $\pm$ 0.13 & 0.06 $\pm$ 0.06 \\
gpt-5-mini & 0.47 $\pm$ 0.02 & 0.63 $\pm$ 0.07 & 0.93 $\pm$ 0.07 & 0.00 $\pm$ 0.00 & 0.03 $\pm$ 0.00 \\
gpt-5 & 0.54 $\pm$ 0.03 & 0.88 $\pm$ 0.04 & 0.99 $\pm$ 0.03 & 0.00 $\pm$ 0.00 & 0.01 $\pm$ 0.04 \\
\midrule
\multicolumn{6}{l}{\textit{Qwen3}}\\
qwen3-14b & 0.19 $\pm$ 0.01 & 0.49 $\pm$ 0.09 & 0.87 $\pm$ 0.03 & 0.00 $\pm$ 0.00 & 0.00 $\pm$ 0.00 \\
qwen3-32b & 0.22 $\pm$ 0.01 & 0.55 $\pm$ 0.12 & 0.91 $\pm$ 0.08 & 0.05 $\pm$ 0.22 & 0.02 $\pm$ 0.04 \\
qwen3-235b-a22b & 0.41 $\pm$ 0.02 & 0.30 $\pm$ 0.00 & 0.73 $\pm$ 0.23 & 1.11 $\pm$ 0.55 & 0.03 $\pm$ 0.06 \\
\midrule
\multicolumn{6}{l}{\textit{GLM}}\\
glm-4-32b & 0.19 $\pm$ 0.01 & 0.47 $\pm$ 0.06 & 0.87 $\pm$ 0.05 & 0.23 $\pm$ 0.16 & 0.54 $\pm$ 0.03 \\
glm-4.5-air & 0.40 $\pm$ 0.01 & 0.47 $\pm$ 0.06 & 0.92 $\pm$ 0.07 & 0.00 $\pm$ 0.00 & 0.09 $\pm$ 0.03 \\
glm-4.5 & 0.44 $\pm$ 0.02 & 0.41 $\pm$ 0.19 & 0.91 $\pm$ 0.08 & 0.03 $\pm$ 0.13 & 0.00 $\pm$ 0.00 \\
\midrule
\multicolumn{6}{l}{\textit{Gemini}}\\
gemini-2.0-flash-001 & 0.46 $\pm$ 0.02 & 0.67 $\pm$ 0.09 & 0.84 $\pm$ 0.00 & 0.00 $\pm$ 0.00 & 0.33 $\pm$ 0.07 \\
gemini-2.5-flash-lite & 0.28 $\pm$ 0.01 & 0.52 $\pm$ 0.04 & 0.91 $\pm$ 0.04 & 0.00 $\pm$ 0.00 & 0.57 $\pm$ 0.02 \\
gemini-2.5-flash & 0.46 $\pm$ 0.02 & 0.66 $\pm$ 0.16 & 0.95 $\pm$ 0.03 & 0.06 $\pm$ 0.13 & 0.59 $\pm$ 0.14 \\
\midrule
\multicolumn{6}{l}{\textit{Claude}}\\
claude-3.5-sonnet & 0.55 $\pm$ 0.31 & 0.82 $\pm$ 0.06 & 1.00 $\pm$ 0.00 & 0.00 $\pm$ 0.00 & 0.39 $\pm$ 0.07 \\
claude-sonnet-4.5 & 0.45 $\pm$ 0.02 & 0.80 $\pm$ 0.00 & 0.95 $\pm$ 0.04 & 0.00 $\pm$ 0.00 & 0.33 $\pm$ 0.00 \\
\midrule
\multicolumn{6}{l}{\textit{Llama}}\\
llama-3.1-70b-instruct & 0.07 $\pm$ 0.04 & 0.63 $\pm$ 0.09 & 0.90 $\pm$ 0.05 & 0.77 $\pm$ 0.56 & 0.47 $\pm$ 0.07 \\
llama-3.1-405b-instruct & 0.18 $\pm$ 0.03 & 0.53 $\pm$ 0.16 & 0.86 $\pm$ 0.09 & 2.29 $\pm$ 0.25 & 0.57 $\pm$ 0.14 \\
\midrule    
\textbf{Average} & 0.37 & 0.57 & 0.91 & 0.28 & 0.35 \\
\bottomrule
\end{tabular}
}
\caption{Psychometric contamination results for PVQ-40 with 95\% confidence intervals from three experimental repetitions.}
\label{tab:pvq}
\end{table*}

\begin{table*}[t]
\centering
{\small
\setlength{\tabcolsep}{3pt}
\renewcommand{\arraystretch}{1.0}
\begin{tabular}{@{}l c c c c c@{}}
\toprule
\textbf{Model} &
\multicolumn{2}{c}{\textbf{Item Memorization}} &
\multicolumn{2}{c}{\textbf{Evaluation Memorization}} &
\multicolumn{1}{c}{\textbf{Target Score Matching}} \\
\cmidrule(lr){2-3} \cmidrule(lr){4-5} \cmidrule(lr){6-6}
& \makecell{\textbf{Semantic}\\\textbf{Memorization}\\\textbf{(Similarity~$\uparrow$)}} &
  \makecell{\textbf{Key Information}\\\textbf{Memorization}\\\textbf{(Success Rate~$\uparrow$)}} &
  \makecell{\textbf{Item--Dimension}\\\textbf{Mapping}\\\textbf{(F1~$\uparrow$)}} &
  \makecell{\textbf{Option--Score}\\\textbf{Mapping}\\\textbf{(MAE~$\downarrow$)}} &
  \makecell{\textbf{(MAE~$\downarrow$)}} \\
\midrule
\multicolumn{6}{l}{\textit{OpenAI}}\\
gpt-4o-mini & 0.21 $\pm$ 0.03 & 0.25 $\pm$ 0.00 & 0.93 $\pm$ 0.07 & 0.27 $\pm$ 0.14 & 0.60 $\pm$ 0.05 \\
gpt-4o & 0.24 $\pm$ 0.03 & 0.35 $\pm$ 0.12 & 1.00 $\pm$ 0.00 & 0.17 $\pm$ 0.38 & 0.73 $\pm$ 0.31 \\
gpt-4.1-nano & 0.19 $\pm$ 0.02 & 0.25 $\pm$ 0.10 & 0.94 $\pm$ 0.08 & 0.70 $\pm$ 0.38 & 1.46 $\pm$ 0.05 \\
gpt-4.1-mini & 0.26 $\pm$ 0.02 & 0.29 $\pm$ 0.10 & 0.87 $\pm$ 0.00 & 0.23 $\pm$ 0.52 & 0.79 $\pm$ 0.16 \\
gpt-4.1 & 0.27 $\pm$ 0.04 & 0.40 $\pm$ 0.06 & 1.00 $\pm$ 0.00 & 0.33 $\pm$ 0.18 & 0.44 $\pm$ 0.13 \\
gpt-5-nano & 0.05 $\pm$ 0.05 & 0.32 $\pm$ 0.12 & 0.97 $\pm$ 0.03 & 0.00 $\pm$ 0.00 & 0.03 $\pm$ 0.05 \\
gpt-5-mini & 0.26 $\pm$ 0.03 & 0.51 $\pm$ 0.06 & 0.99 $\pm$ 0.05 & 0.00 $\pm$ 0.00 & 0.01 $\pm$ 0.03 \\
gpt-5 & 0.70 $\pm$ 0.16 & 0.61 $\pm$ 0.06 & 1.00 $\pm$ 0.00 & 0.00 $\pm$ 0.00 & 0.00 $\pm$ 0.00 \\
\midrule
\multicolumn{6}{l}{\textit{Qwen3}}\\
qwen3-14b & 0.24 $\pm$ 1.39 & 0.22 $\pm$ 0.16 & 0.94 $\pm$ 0.05 & 0.07 $\pm$ 0.16 & 0.16 $\pm$ 0.20 \\
qwen3-32b & 0.19 $\pm$ 0.03 & 0.22 $\pm$ 0.06 & 0.99 $\pm$ 0.05 & 0.34 $\pm$ 0.39 & 0.12 $\pm$ 0.12 \\
qwen3-235b-a22b & 0.29 $\pm$ 0.05 & 0.17 $\pm$ 0.00 & 0.75 $\pm$ 0.17 & 0.86 $\pm$ 0.63 & 0.00 $\pm$ 0.00 \\
\midrule
\multicolumn{6}{l}{\textit{GLM}}\\
glm-4-32b & 0.21 $\pm$ 0.02 & 0.17 $\pm$ 0.00 & 1.00 $\pm$ 0.00 & 0.18 $\pm$ 0.43 & 0.60 $\pm$ 0.11 \\
glm-4.5-air & 0.29 $\pm$ 0.04 & 0.25 $\pm$ 0.10 & 0.88 $\pm$ 0.05 & 0.07 $\pm$ 0.29 & 0.10 $\pm$ 0.11 \\
glm-4.5 & 0.27 $\pm$ 0.03 & 0.21 $\pm$ 0.10 & 0.99 $\pm$ 0.05 & 0.00 $\pm$ 0.00 & 0.00 $\pm$ 0.00 \\
\midrule
\multicolumn{6}{l}{\textit{Gemini}}\\
gemini-2.0-flash-001 & 0.22 $\pm$ 0.03 & 0.22 $\pm$ 0.06 & 0.64 $\pm$ 0.15 & 0.20 $\pm$ 0.00 & 0.50 $\pm$ 0.17 \\
gemini-2.5-flash-lite & 0.19 $\pm$ 0.03 & 0.24 $\pm$ 0.12 & 0.96 $\pm$ 0.00 & 0.40 $\pm$ 0.00 & 0.73 $\pm$ 0.16 \\
gemini-2.5-flash & 0.20 $\pm$ 0.03 & 0.28 $\pm$ 0.06 & 0.95 $\pm$ 0.05 & 0.10 $\pm$ 0.00 & 0.49 $\pm$ 0.04 \\
\midrule
\multicolumn{6}{l}{\textit{Claude}}\\
claude-3.5-sonnet & 0.21 $\pm$ 0.14 & 0.51 $\pm$ 0.12 & 1.00 $\pm$ 0.00 & 0.00 $\pm$ 0.00 & 0.11 $\pm$ 0.04 \\
claude-sonnet-4.5 & 0.11 $\pm$ 0.00 & 0.61 $\pm$ 0.06 & 0.95 $\pm$ 0.00 & 0.00 $\pm$ 0.00 & 0.66 $\pm$ 0.02 \\
\midrule
\multicolumn{6}{l}{\textit{Llama}}\\
llama-3.1-70b-instruct & 0.05 $\pm$ 0.02 & 0.28 $\pm$ 0.16 & 0.86 $\pm$ 0.04 & 1.67 $\pm$ 0.04 & 0.67 $\pm$ 0.12 \\
llama-3.1-405b-instruct & 0.10 $\pm$ 0.08 & 0.18 $\pm$ 0.06 & 0.86 $\pm$ 0.08 & 2.23 $\pm$ 1.09 & 0.37 $\pm$ 0.09 \\
\midrule
\textbf{Average} & 0.24 & 0.31 & 0.93 & 0.37 & 0.41 \\
\bottomrule
\end{tabular}
}
\caption{Psychometric contamination results for SD-3 with 95\% confidence intervals from three experimental repetitions.}
\label{tab:sd3}
\end{table*}

\begin{table*}[htbp]
\centering
{\small
\setlength{\tabcolsep}{6pt}
\renewcommand{\arraystretch}{1.05}
\begin{tabular}{@{}l c c@{}}
\toprule
\textbf{Model} &
\makecell{\textbf{Refusal Rate (\%)}} &
\makecell{\textbf{Similarity Score}\\\textbf{(6-point scale)}} \\
\midrule

\multicolumn{3}{l}{\textit{OpenAI}}\\
gpt-4o-mini      & 0.0 $\pm$ 1.4   & 2.02 $\pm$ 0.28 \\
gpt-4o           & 12.9 $\pm$ 5.7  & 1.77 $\pm$ 0.28 \\
gpt-4.1-nano     & 9.1 $\pm$ 5.0   & 1.68 $\pm$ 0.25 \\
gpt-4.1-mini     & 0.0 $\pm$ 1.4   & 1.72 $\pm$ 0.25 \\
gpt-4.1          & 0.0 $\pm$ 1.4   & 1.92 $\pm$ 0.29 \\
gpt-5-nano       & 62.3 $\pm$ 8.2  & 1.90 $\pm$ 0.51 \\
gpt-5-mini       & 1.5 $\pm$ 2.5   & 3.07 $\pm$ 0.40 \\
gpt-5            & 14.4 $\pm$ 6.0  & 5.74 $\pm$ 0.16 \\
\midrule
\addlinespace[3pt]

\multicolumn{3}{l}{\textit{Qwen3}}\\
qwen3-14b        & 15.2 $\pm$ 6.1  & 1.59 $\pm$ 0.25 \\
qwen3-32b        & 2.3 $\pm$ 2.8   & 1.52 $\pm$ 0.20 \\
qwen3-235b-a22b  & 0.0 $\pm$ 1.8   & 1.69 $\pm$ 0.29 \\
\midrule
\addlinespace[3pt]

\multicolumn{3}{l}{\textit{GLM}}\\
glm-4-32b        & 6.1 $\pm$ 4.2   & 1.66 $\pm$ 0.25 \\
glm-4.5-air      & 2.3 $\pm$ 2.9   & 1.86 $\pm$ 0.26 \\
glm-4.5          & 0.0 $\pm$ 1.4   & 3.32 $\pm$ 0.39 \\
\midrule
\addlinespace[3pt]

\multicolumn{3}{l}{\textit{Gemini}}\\
gemini-2.0-flash-001   & 0.0 $\pm$ 1.4  & 1.64 $\pm$ 0.24 \\
gemini-2.5-flash-lite  & 4.5 $\pm$ 3.7  & 1.67 $\pm$ 0.25 \\
gemini-2.5-flash       & 0.0 $\pm$ 1.4  & 1.89 $\pm$ 0.30 \\
\midrule
\addlinespace[3pt]

\multicolumn{3}{l}{\textit{Claude}}\\
claude-3.5-sonnet & 98.5 $\pm$ 2.5 & 1.50 $\pm$ 6.35 \\
claude-4.5-sonnet & 71.2 $\pm$ 7.6 & 4.63 $\pm$ 0.65 \\
\midrule
\addlinespace[3pt]

\multicolumn{3}{l}{\textit{Llama}}\\
llama-3.1-70b-instruct   & 87.9 $\pm$ 5.6 & 1.56 $\pm$ 0.82 \\
llama-3.1-405b-instruct  & 97.7 $\pm$ 2.8 & 1.00 $\pm$ 0.00 \\
\bottomrule
\end{tabular}
}
\caption{The refusal rate of each model and the semantic similarity score with 95\% confidence intervals on BFI-44.}
\label{tab:bfi44_refusal_ci}
\end{table*}

\begin{table*}[htbp]
\centering
{\small
\setlength{\tabcolsep}{6pt}
\renewcommand{\arraystretch}{1.05}
\begin{tabular}{@{}l c c@{}}
\toprule
\textbf{Model} &
\makecell{\textbf{Refusal Rate (\%)}} &
\makecell{\textbf{Similarity Score}\\\textbf{(6-point scale)}} \\
\midrule

\multicolumn{3}{l}{\textit{OpenAI}}\\
gpt-4o-mini      & 1.0 $\pm$ 2.7   & 1.19 $\pm$ 0.17 \\
gpt-4o           & 99.0 $\pm$ 2.7  & 2.00 $\pm$ 0.00 \\
gpt-4.1-nano     & 3.1 $\pm$ 3.9   & 1.29 $\pm$ 0.21 \\
gpt-4.1-mini     & 0.0 $\pm$ 1.9   & 1.06 $\pm$ 0.09 \\
gpt-4.1          & 0.0 $\pm$ 1.9   & 2.06 $\pm$ 0.34 \\
gpt-5-nano       & 70.8 $\pm$ 9.0  & 1.50 $\pm$ 0.57 \\
gpt-5-mini       & 2.1 $\pm$ 3.4   & 1.68 $\pm$ 0.32 \\
gpt-5            & 41.1 $\pm$ 9.7  & 2.12 $\pm$ 0.49 \\
\midrule
\addlinespace[3pt]

\multicolumn{3}{l}{\textit{Qwen3}}\\
qwen3-14b        & 26.0 $\pm$ 8.7  & 1.24 $\pm$ 0.23 \\
qwen3-32b        & 5.2 $\pm$ 4.7   & 1.25 $\pm$ 0.20 \\
qwen3-235b-a22b  & 0.0 $\pm$ 2.6   & 1.99 $\pm$ 0.36 \\
\midrule

\multicolumn{3}{l}{\textit{GLM}}\\
glm-4-32b        & 0.0 $\pm$ 1.9   & 1.23 $\pm$ 0.17 \\
glm-4.5-air      & 20.5 $\pm$ 8.9  & 1.16 $\pm$ 0.21 \\
glm-4.5          & 6.2 $\pm$ 5.0   & 1.43 $\pm$ 0.25 \\
\midrule

\multicolumn{3}{l}{\textit{Gemini}}\\
gemini-2.0-flash-001   & 0.0 $\pm$ 1.9  & 1.19 $\pm$ 0.19 \\
gemini-2.5-flash-lite  & 40.6 $\pm$ 9.6 & 1.21 $\pm$ 0.21 \\
gemini-2.5-flash       & 0.0 $\pm$ 1.9  & 1.43 $\pm$ 0.20 \\
\midrule

\multicolumn{3}{l}{\textit{Claude}}\\
claude-3.5-sonnet & 97.9 $\pm$ 3.4 & 1.00 $\pm$ 0.00 \\
claude-4.5-sonnet & 96.9 $\pm$ 3.9 & 6.00 $\pm$ 0.00 \\
\midrule

\multicolumn{3}{l}{\textit{Llama}}\\
llama-3.1-70b-instruct   & 77.1 $\pm$ 8.3 & 1.36 $\pm$ 0.54 \\
llama-3.1-405b-instruct  & 96.9 $\pm$ 3.9 & 1.00 $\pm$ 0.00 \\
\bottomrule
\end{tabular}
}
\caption{The refusal rate of each model and the semantic similarity score with 95\% confidence intervals on MFQ.}
\label{tab:mfq_refusal_ci}
\end{table*}

\begin{table*}[htbp]
\centering
{\small
\setlength{\tabcolsep}{6pt}
\renewcommand{\arraystretch}{1.05}
\begin{tabular}{@{}l c c@{}}
\toprule
\textbf{Model} &
\makecell{\textbf{Refusal Rate (\%)}} &
\makecell{\textbf{Similarity Score}\\\textbf{(6-point scale)}} \\
\midrule

\multicolumn{3}{l}{\textit{OpenAI}}\\
gpt-4o-mini      & 31.7 $\pm$ 5.8  & 1.55 $\pm$ 0.20 \\
gpt-4o           & 42.5 $\pm$ 6.2  & 1.93 $\pm$ 0.27 \\
gpt-4.1-nano     & 18.3 $\pm$ 4.9  & 1.56 $\pm$ 0.17 \\
gpt-4.1-mini     & 0.0 $\pm$ 0.8   & 1.70 $\pm$ 0.18 \\
gpt-4.1          & 0.0 $\pm$ 0.8   & 1.86 $\pm$ 0.20 \\
gpt-5-nano       & 92.9 $\pm$ 3.3  & 1.71 $\pm$ 0.67 \\
gpt-5-mini       & 9.6 $\pm$ 3.7   & 1.83 $\pm$ 0.19 \\
gpt-5            & 16.0 $\pm$ 4.6  & 2.77 $\pm$ 0.29 \\
\midrule

\multicolumn{3}{l}{\textit{Qwen3}}\\
qwen3-14b        & 15.1 $\pm$ 4.5  & 1.18 $\pm$ 0.10 \\
qwen3-32b        & 19.2 $\pm$ 5.0  & 1.38 $\pm$ 0.15 \\
qwen3-235b-a22b  & 0.0 $\pm$ 0.9   & 1.88 $\pm$ 0.22 \\
\midrule

\multicolumn{3}{l}{\textit{GLM}}\\
glm-4-32b        & 3.8 $\pm$ 2.5   & 1.50 $\pm$ 0.15 \\
glm-4.5-air      & 0.9 $\pm$ 1.5   & 1.61 $\pm$ 0.17 \\
glm-4.5          & 0.4 $\pm$ 1.2   & 1.88 $\pm$ 0.21 \\
\midrule

\multicolumn{3}{l}{\textit{Gemini}}\\
gemini-2.0-flash-001   & 0.0 $\pm$ 0.8  & 1.74 $\pm$ 0.18 \\
gemini-2.5-flash-lite  & 0.8 $\pm$ 1.4  & 1.49 $\pm$ 0.15 \\
gemini-2.5-flash       & 0.0 $\pm$ 0.8  & 1.78 $\pm$ 0.18 \\
\midrule

\multicolumn{3}{l}{\textit{Claude}}\\
claude-3.5-sonnet & 97.9 $\pm$ 1.9 & 2.00 $\pm$ 2.78 \\
claude-4.5-sonnet & 0.0 $\pm$ 0.8  & 1.98 $\pm$ 0.21 \\
\midrule

\multicolumn{3}{l}{\textit{Llama}}\\
llama-3.1-70b-instruct   & 92.9 $\pm$ 3.3 & 1.00 $\pm$ 0.00 \\
llama-3.1-405b-instruct  & 88.3 $\pm$ 4.1 & 1.21 $\pm$ 0.31 \\
\bottomrule
\end{tabular}
}
\caption{The refusal rate of each model and the semantic similarity score with 95\% confidence intervals on PVQ-40.}
\label{tab:pvq40_refusal_ci}
\end{table*}

\begin{table*}[htbp]
\centering
{\small
\setlength{\tabcolsep}{6pt}
\renewcommand{\arraystretch}{1.05}
\begin{tabular}{@{}l c c@{}}
\toprule
\textbf{Model} &
\makecell{\textbf{Refusal Rate}} &
\makecell{\textbf{Similarity Score}\\\textbf{(6-point scale)}} \\
\midrule

\multicolumn{3}{l}{\textit{OpenAI}}\\
gpt-4o-mini      & 0.0 $\pm$ 2.5   & 1.32 $\pm$ 0.23 \\
gpt-4o           & 36.1 $\pm$ 10.8 & 1.54 $\pm$ 0.36 \\
gpt-4.1-nano     & 12.5 $\pm$ 7.7  & 1.38 $\pm$ 0.25 \\
gpt-4.1-mini     & 1.4 $\pm$ 3.6   & 1.90 $\pm$ 0.32 \\
gpt-4.1          & 0.0 $\pm$ 2.5   & 2.22 $\pm$ 0.40 \\
gpt-5-nano       & 91.7 $\pm$ 6.6  & 1.00 $\pm$ 0.00 \\
gpt-5-mini       & 23.6 $\pm$ 9.7  & 2.60 $\pm$ 0.49 \\
gpt-5            & 69.4 $\pm$ 10.4 & 4.18 $\pm$ 1.06 \\
\midrule

\multicolumn{3}{l}{\textit{Qwen3}}\\
qwen3-14b        & 97.2 $\pm$ 4.4  & 3.00 $\pm$ 25.41 \\
qwen3-32b        & 34.7 $\pm$ 10.7 & 1.53 $\pm$ 0.37 \\
qwen3-235b-a22b  & 1.8 $\pm$ 4.6   & 2.47 $\pm$ 0.54 \\
\midrule

\multicolumn{3}{l}{\textit{GLM}}\\
glm-4-32b        & 0.0 $\pm$ 2.5   & 1.53 $\pm$ 0.25 \\
glm-4.5-air      & 19.4 $\pm$ 9.7  & 2.08 $\pm$ 0.47 \\
glm-4.5          & 0.0 $\pm$ 2.5   & 2.17 $\pm$ 0.38 \\
\midrule

\multicolumn{3}{l}{\textit{Gemini}}\\
gemini-2.0-flash-001   & 0.0 $\pm$ 2.5  & 1.88 $\pm$ 0.37 \\
gemini-2.5-flash-lite  & 40.3 $\pm$ 11.0 & 1.30 $\pm$ 0.27 \\
gemini-2.5-flash       & 1.4 $\pm$ 3.6  & 1.45 $\pm$ 0.26 \\
\midrule

\multicolumn{3}{l}{\textit{Claude}}\\
claude-3.5-sonnet & 93.1 $\pm$ 6.1 & 1.60 $\pm$ 1.67 \\
claude-4.5-sonnet & 98.6 $\pm$ 3.6 & 1.00 $\pm$ 0.00 \\
\midrule

\multicolumn{3}{l}{\textit{Llama}}\\
llama-3.1-70b-instruct   & 69.4 $\pm$ 10.4 & 1.00 $\pm$ 0.00 \\
llama-3.1-405b-instruct  & 95.8 $\pm$ 5.1  & 1.00 $\pm$ 0.00 \\
\bottomrule
\end{tabular}
}
\caption{The refusal rate of each model and the semantic similarity score with 95\% confidence intervals on SD-3.}
\label{tab:sd3_refusal_ci}
\end{table*}

\section{Checklist for Responsible NLP Research}
\subsection{The License of Artifacts}
We used the official questionnaire items from the BFI-44, MFQ, PVQ-40, and SD-3 inventories. All of these instruments are publicly available for non-commercial academic research and can be freely used for scholarly purposes.
\subsection{Artifact Use Consistent with Intended Use}
We followed the original administration and scoring procedures of each inventory as described in their source publications, ensuring that the use of all questionnaire items remained consistent with their intended psychometric purpose.
\subsection{Documentation of Artifacts}
All inventories are in English.
\subsection{Statistics for Data}
The BFI-44 consists of 44 items, the MFQ includes 32 items (of which 2 are attention checks), the PVQ-40 contains 40 items, and the SD-3 comprises 24 items.
\subsection{Descriptive Statistics}
All experiments were conducted three times with the temperature set to 0.7. We report the mean along with 95\% intervals.
\subsection{AI Assistants in Research or Writing}
We used AI assistants to refine writing, proofread the text, and assist with coding experiments.

\end{document}